\documentclass{article}
\usepackage{ijcai18}

\usepackage{times}
\usepackage{xcolor}
\usepackage{soul}
\usepackage[utf8]{inputenc}
\usepackage[small]{caption}

\usepackage{amsmath}
\usepackage{subfigure}
\usepackage[pdftex]{graphicx}

\usepackage{xcolor}

\title{
  Combining Linear Non-Gaussian Acyclic Model with Logistic Regression Model\\
  for Estimating Causal Structure from Mixed Continuous and Discrete Data
}

\author{
Chao Li$^1$,
Shohei Shimizu$^{1,2}$
\\
$^1$ RIKEN AIP Center, Japan\\
$^2$ Shiga University, Japan\\
chao.li.hf@riken.jp,
shohei-shimizu@biwako.shiga-u.ac.jp
}

\begin{document}

\maketitle

\begin{abstract}
Estimating causal models from observational data is a crucial task in data analysis.
For continuous-valued data, Shimizu et al.\ have proposed a linear acyclic non-Gaussian model to understand the data generating process, and have shown that their model is identifiable when the number of data is sufficiently large.
However, situations in which continuous and discrete variables coexist in the same problem are common in practice.
Most existing causal discovery methods either ignore the discrete data and apply a continuous-valued  algorithm or discretize all the continuous data and then apply a discrete Bayesian network approach.
These methods possibly loss important information when we ignore discrete data or introduce the approximation error due to discretization.
In this paper, we define a novel hybrid causal model which consists of both continuous and discrete variables.
The model assumes: (1) the value of a continuous variable is a linear function of its parent variables plus a non-Gaussian noise, and (2) each discrete variable is a logistic variable whose distribution parameters depend on the values of its parent variables.
In addition, we derive the BIC scoring function for model selection.
The new discovery algorithm can learn causal structures from mixed continuous and discrete data without discretization.
We empirically demonstrate the power of our method through thorough simulations.
\end{abstract}

\section{Introduction}
Estimating a causal directed acyclic graph (DAG) model (also known as causal Bayesian network) from observational data is a challenging problem and has applications in many research areas, including bioinformatics, economics and social science~\cite{Spirtes1993,Pearl2000,Morgan2007}.
Existing methods for causal discovery commonly assume the involved variables are either discrete, or continuous valued.
In the discrete case, one of the principled approaches is the constrained-based method, which relies on the results of conditional independence tests.
While this approach dose not impose any functional assumptions on the dependencies, it can not identify the causal directions from two different DAG models that entail identical set of conditional independencies.
To cope with this identifiability problem, \cite{Peters11} extend the additive noise model to the discrete case and demonstrate the causal direction in their model can be identified in general.
However, the model imposes linear dependence assumptions on the data and this assumption is not often satisfied in practice, especially for binary categorical data.
For the multivariate count data, \cite{ParkR15} proposed Poisson DAG model in which each node corresponds to a Poisson random variable with rate parameters depending only on its parent variables.
Again, the model can be applied on only count data instead of categorical data.

For the continuous-valued data, the traditional methods for causal discovery are based on linear model with Gaussian noise~\cite{Geiger1994,Spirtes1993}.
However, the linear Gaussian approach usually outputs a set of possible models which belong to the Markov equivalence class of the true model.
To avoid this limitation, \citeauthor{shimizu06a}~\shortcite{shimizu06a} proposed linear non-Gaussian acyclic model (LiNGAM) and showed that the full causal structure is identifiable given sufficiently large number of data.
To relax the assumption of all variables are non-Gaussian, \cite{Hoyer08}  proposed the PClingam algorithm which combines the independence based PC algorithm~\cite{Spirtes1993} and the ICA-based LiNGAM algorithm~\cite{shimizu06a}.
The algorithm first use the PC algorithm to obtain a set of candidate DAG models, and then apply a scoring directions scheme for model selection.

While real data often contains a mixture of discrete and continuous variables, all approaches so far we described have assumed data are either discrete or continuous.
One of the commonly employed approach for mixed data is to ignore the discrete variable and apply a linear causal network approach for only continuous data. The causal analysis losses sight of some important information due to ignorance of discrete data.
Another one is to discretize the continuous variables and then apply the discrete Bayesian network to analysis causal relationships, since many efficient Bayesian network learning algorithm~\cite{Spirtes1993,Yehezkel2009,YuanM13} has been proposed for discrete data.
The choice of discretization policy has significant impact on the resulting model and the discretization may lead to wrong model if much information is lost due to discretization process.
Recently, \citeauthor{Chen2017}~\shortcite{Chen2017} proposed a new discretization strategy to mitigate these problem.
Nevertheless, these traditional methods learn a Markov equivalence class and therefore the causal directions of some edges can not be determined.

In this contribution, we propose a novel hybrid causal model which consists of both continuous and discrete variables.
Our model is based on the LiNGAM model and logistic regression model.
In our model, we assume:
\begin{enumerate}
  \item The data generating process can be represented by a directed acyclic graph.

  \item Each continuous variable is generated from a linear function of its parent variables plus a non-Gaussian noise.
   
  \item Each discrete variable is a logistic variable which depends on its parent variables.
\end{enumerate}
An important features of this model is that the model can handle continuous and discrete variables simultaneously without using discretization.
In addition, we derive the BIC scoring function for evaluating possible model and we also propose to use the BIC score for causal discovery.
Most constraint-based discovery algorithms, e.g.~the PC algorithm~\cite{Spirtes1993}, find a Markov equivalent class which is a set of DAG models.
In contrast, our method leverage the identifiability of the LiNGAM model, are expected to be able to identify the full causal structure from observational data.
Finally, we empirically demonstrate the power of our method through thorough simulations.

The remainder of this paper is structured as follows:
Section~\ref{sec:back} summaries the necessary notation and reviews the LiNGAM model and logistic model.
Section~\ref{sec:hybrid} defines our hybrid causal model and derives the BIC scoring function for model selection.
Section \ref{sec:exp} empirically evaluates our methods. 
Section \ref{sec:conclude} concludes this paper.

\section{Background}  \label{sec:back}
In this section, we first introduce some necessary notation and definitions for directed acyclic graph (DAG) models.
Then we briefly review the two building blocks of our model: the linear non-Gaussian acyclic model (LiNGAM) and 
logistic conditional probability distribution.

\subsection{DAG Models}

Let us consider a set of random variables $\mathbf{X} = \{X_1, X_2, \ldots, X_p\}$ with index set $\mathbf{V}=\{1, 2, \ldots, p\}$.
Following the convention of previous studies,
a causal graph over a set of variables $\mathbf{X}$ is a DAG $G = (\mathbf{V}, \mathbf{E})$ with node set $\mathbf{V}=\{1, 2, \ldots, p\}$, which represents the random variables $\mathbf{X}=\{X_1, X_2, \ldots, X_p\}$, and edge set $\mathbf{E}$ (or lack of them), which represents direct dependency relationships (or conditional independence relationships) between variables.
A directed edge from node $i$ to node $j$ is denoted by $(i,j)$ or $i \rightarrow j$.
A node $i$ is called a parent of $j$ if $(i,j) \in \mathbf{E}$ and the parent set $\mathbf{PA}(i)$ of a node $i$ consists of all nodes $j$ such that $(j,i) \in \mathbf{E}$.
The joint probability distribution $p(\mathbf{X})$ of variables $\mathbf{X}$ can be factorized in terms of the conditional probability distributions as follows:
\begin{equation}
P(\mathbf{X}) = P(X_1, X_2, \ldots, X_p) = \prod_{i=1}^{p} P(X_i\mid X_{\mathbf{PA}(i)}),
\end{equation}
where $P(X_i\mid X_{\mathbf{PA}(i)})$ refers to the conditional probability distribution of $X_i$ given its parents variables $X_{\mathbf{PA}(i)}$.
For the variables $X_i$ without parents (called root variables), $P(X_i\mid X_{\mathbf{PA}(i)})$ stands for marginal distribution $P(X_i)$.

The set of all independence constraints imposed by the structure of a DAG model can be characterized by the Markov conditions, which are the constraints that each variable is independent of its non-descendants given its parents.
Two DAG structures are Markov equivalent if the set of conditional independence constraints imposed by one DAG is identical to that of another DAG.
A Markov equivalence class is a set of DAGs that encode the same set of conditional independencies.
The constraint based causal discovery algorithm requires a faithfulness assumption: the conditional independencies in the data distribution exactly equal the ones encoded in the causal structure.
Because the constraint based approach to causal inference considers only independence constraints, these methods find a Markov equivalent class of the true causal structure.

To identify more edge directions of estimated causal structure, we propose to combine the LiNGAM model and logistic model for causal discovery.
Therefore, we review these two concepts in the next two subsections.

\begin{figure}[!ht]
  \begin{center}
    \includegraphics[width=0.45\textwidth]{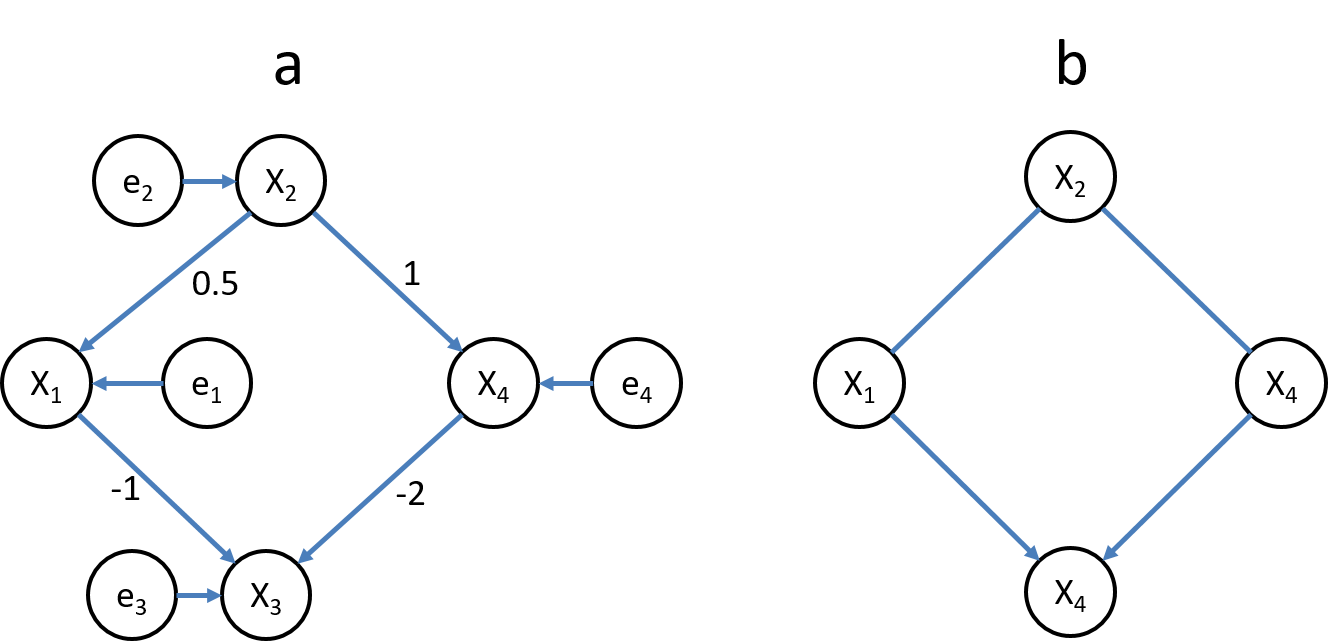}
  \end{center}
  \caption{ (a) An example of the LiNGAM model, and (b) the Markov equivalence class of the DAG on the left.}
  \label{fig:lingam}
\end{figure}

\subsection{LiNGAM}

To estimate a causal structure from continuous data, \citeauthor{shimizu06a}~\shortcite{shimizu06a} proposed a linear non-Gaussian acyclic model (LiNGAM), which is a special case of structural equation models and continuous-valued Bayesian networks.
The LiNGAM model assumes that the observed data are generated from a process which is represented graphically by a directed acyclic graph (DAG).
Moreover, it assumes that the relations between the variables are linear.
Let us denote a connection strength from a variable $x_j$ to another variable $x_i$ in the DAG by $b_{ij}$,
then the model can be represented by 
\begin{equation}
X_i = e_i + b_{i0} +  \sum_{j \in \mathbf{PA}(i) }{b_{ij} X_j}  \ with\ e_i \sim \operatorname{non-Gaussian}
\end{equation}
where $e_i$ is called an noise variable.
All noise variables $e_i$ are continuous random variables having non-Gaussian distributions with zero means and non-zero variances, and $e_i$ are independent of each other so that there are no latent confounding variables~\cite{Spirtes1993}.
See Figure~\ref{fig:lingam}a for a concrete example of LiNGAM model, the data is generated by first drawing the $e_i$ independently from their respective non-Gaussian distributions, and subsequently setting (in an topological order) $X_2 = e_2$, $X_1 = 0.5\times X_2 + e_1$, $X_4 = X_2 +e_4$, and $X_3 = -1 \times X_1 -2 \times X_4 + e_3$.
(Here, we have assumed for simplicity that all the $b_{i0}$ are zero, but this may not be the case in general.) 
A remarkable result was shown in \citeauthor{shimizu06a}~\shortcite{shimizu06a} is that under the non-Gaussian assumption about the noise distribution, the full causal structure and associated parameters are identifiable.
In contrast, the constraint-based algorithms estimate the Markov equivalence class and thus the directions of some edges can not be estimated (see Figure~\ref{fig:lingam}b).

\subsection{Logistic Conditional Probability Distribution}

In this section, we describe the logistic regression model as a local causal structure.
Consider a discrete variable $Y$ whose distribution depends on some set of causes $X_1, X_2, \ldots, X_k$.
In this study, we restrict our analysis on binary variables for $Y$ which takes two values $\{1, 2\}$.
We assume that the conditional probability distribution of $Y$ given its dependent variables is a logistic CPD, which is defined as follows.

Let $Y$ be a binary-valued random variable defined over the domain $\{1, 2\}$, with $k$ parents $X_1, X_2, \ldots, X_k$ that take on numerical values.
The conditional probability distribution (CPD) $P(Y \mid X_1,X_2,X_k)$  of $Y$ is a logistic CPD if there are $k+1$ weights $b_{0}, b_1, \ldots, b_k$ such that:
\begin{equation}
P(Y=1 \mid X_1,X_2, \ldots, X_k) = sigmoid(b_{0} + \sum_{j =1}^{k}{b_{j} X_j)},
\end{equation}
where the sigmoid function stands for:
\begin{equation}
sigmoid(z) = \frac{e^z}{1 + e^z}
\end{equation}

The logistic CPD is a natural model for many real-world applications, because it naturally aggregates the influence of different parents.
\citeauthor{Koller+Friedman:09}~\shortcite{Koller+Friedman:09} also provide a variant of binary logistic CPD which can handle the multi-valued variables, however, we have not implemented this feature in our software, yet.

\section{The Hybrid Causal Model}  \label{sec:hybrid}

In this section, we propose a novel hybrid causal model, i.e., a DAG model consisting of both continuous and discrete variables.
Then we discuss two commonly used approaches to causal discovery.
Finally, we derive the BIC scoring function to evaluate the fitness of a DAG model to the data.

\subsection{Definition of Our Model}

We partition the variables of our model in two types: continuous variables and discrete variables.
In this paper, we assume that each discrete variable only take two values $\{1,2\}$ and assume that the observed data has been generated by the following process:

\begin{enumerate}
  \item The data are generated from a process represented graphically by a directed acyclic graph, in which each variable is directly caused by its parent variables.
  
  \item The value assigned to each continuous variables $X_i$ is a linear function of its parent variables plus a non-Gaussian noise term $e_i$, that is 
\begin{equation}
X_i = e_i + b_{i0} + \sum_{j \in \mathbf{PA}(i) }{b_{ij} X_j},
\end{equation}
where the noise term $e_i$ are all continuous random variables with non-Gaussian densities, and the noise variables $e_i$ are independent of each other.
  
  \item For a discrete variable $X_i$, the conditional probability distribution of variable $X_i$ is the logistic CPD, such that,
\begin{equation}
P(X_i = 1 \mid X_{\mathbf{PA}(i)} ) = sigmoid(b_{0i} + \sum_{j \in \mathbf{PA}(i) }{b_{ij} X_j)},
\end{equation}
\begin{equation}
P(X_i = 2 \mid X_{\mathbf{PA}(i)} ) = 1 - P(X_i = 1 \mid X_{\mathbf{PA}(i)} ),
\end{equation}

\end{enumerate}
  

\subsection{Discovery Algorithms}

Causal discovery consists in finding the causal model that best fits the sample data according to certain criterion.
Since a causal model consists of a causal graph structure (a DAG) and associated parameters, the discovery algorithms often need to deal with two highly related tasks: search for a causal graph and estimation of the parameters.
That is, in order to estimate the parameters, we must know the causal structure; in order to evaluate a candidate causal structure, we must estimate the parameters from the data and the causal graph.
In this paper, we are mainly interested in algorithms for learning the causal structure, and view the parameter estimation part as a subroutine of the search algorithm.

The constraint-based algorithms typically apply statistical tests to identify conditional independence relations and attempt to find a causal graph that represents these relations as precise as possible.
Since the accuracy of the statistical test is sensitive to the number of data and the complexity of the independence tests, the constraint-based algorithms may not work well when there are insufficient data.
Another issue in the constraint-based algorithm is that independence test based approach can not distinguish two DAGs in the same Markov equivalence class.
Since most of Markov equivalence classes contain more than one graph, conditional independence based methods leave some arrows undirected and cannot uniquely identify the true causal graph.
Recently, \citeauthor{Hoyer08}~\shortcite{Hoyer08} use constraint-based methods to infer the Markov equivalence class of the true causal model and then score each DAG belonging to the equivalence class.

\subsection{Scoring the Hybrid Causal Model}

In order to evaluate the hybrid causal model, we derive the Bayesian information criterion (BIC) scoring function ~\cite{schwarz1978} of our model.
The basic idea of the BIC is to select the causal structure that maximizes the log-likelihood and being penalized by the number of parameters which are necessary to specify the causal model.

Let us denote our model as a pair $\langle G, B_G \rangle$, where $G$ denotes the causal graph and $B_G$ represents all parameter for the graph structure $G$.
The BIC score of a structure $G$ can be defined as:
\begin{equation}  \label{eq:bic}
Score_{BIC}(G) = log\mathcal{L}(\hat{B_G}:D) - \frac{log(M)}{2}Dim[G],
\end{equation}
where $log\mathcal{L}(\hat{B_G}:D)$ is the logarithm of the likelihood function, $\hat{B_G}$ are the maximum likelihood estimated (MLE) parameters for $G$, $M$ stands for the number of data points, and $Dim[G]$ signifies the number of free parameters of the model.

Since in our model, each conditional probability distribution $P(X_i\mid X_{\mathbf{PA}(i)})$ have the number of its parent variables plus one constant parameter, the total number of parameters $Dim[G]$ is the number of edges plus the number of variables.

Next, we discuss the methods for obtain the MLE parameters.
For the logistic variables, it is easy to estimate dependence coefficients by the maximum likelihood principle~\cite{Bishop:2006}.
however, for the continuous variables, as we do not assume the Gaussian noise, to obtain the exact MLE parameters, we have to estimate the noise distribution first.
This would be very complicated and not easy for implementation.
For simplicity of the estimation, we obtain the coefficients $b_{ij}$ estimated using ordinary least-square regression. Note this provide a consistent estimates. 
When the number of data is sufficiently large, the approximation error becomes zero.

Finaly, for the continuous variable $X_i$, the local log-likelihood $logL(X_i)$ of the variable $X_i$ is given by \citeauthor{Aapo2010}~\shortcite{Aapo2010}. 
For the discrete variable, we just use canonical likelihood function for logistic regression~\cite{Bishop:2006}.

For the discrete Bayesian network, it is well known that the BIC scoring function assigns the same score to structures in the same equivalence class.
However, under assumptions described in subsection 3.1, using BIC scoring function we are expected to be able to find the unique true causal structure as well as the associated parameters.

\section{Experiments}  \label{sec:exp}

In this section, we evaluated our proposed algorithm with respect to accuracy rate of discovering causal structure through extensive experiments.
We also showed that our algorithm has better performance compared to the state of the art algorithm.

\subsection{Simulations}

\begin{figure*}[!tb]
  \centering
  \subfigure[c=1]{%
    \includegraphics[width=0.33\textwidth]{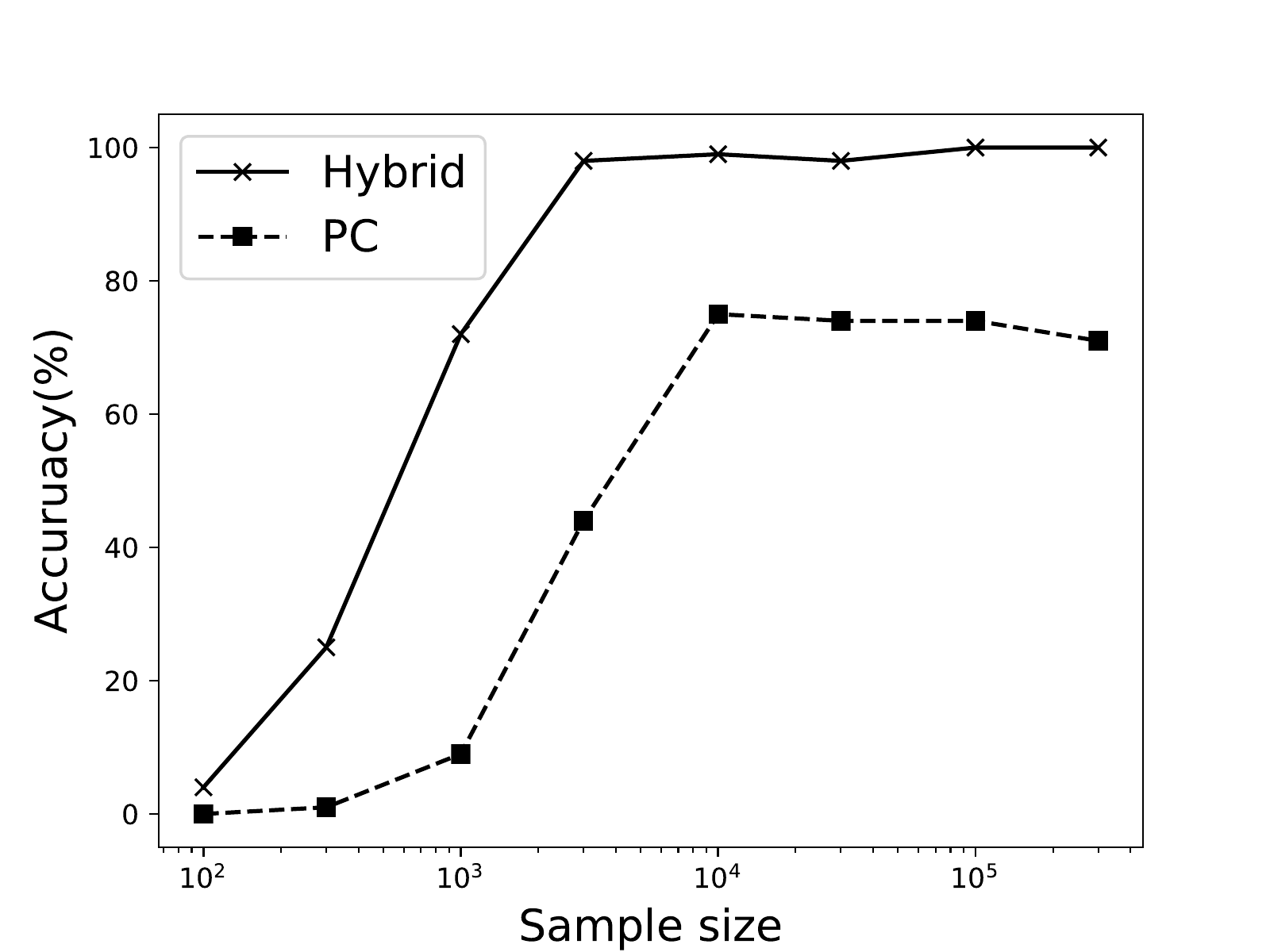}%
  }%
  \hfill
  \subfigure[c=2]{%
    \includegraphics[width=0.33\textwidth]{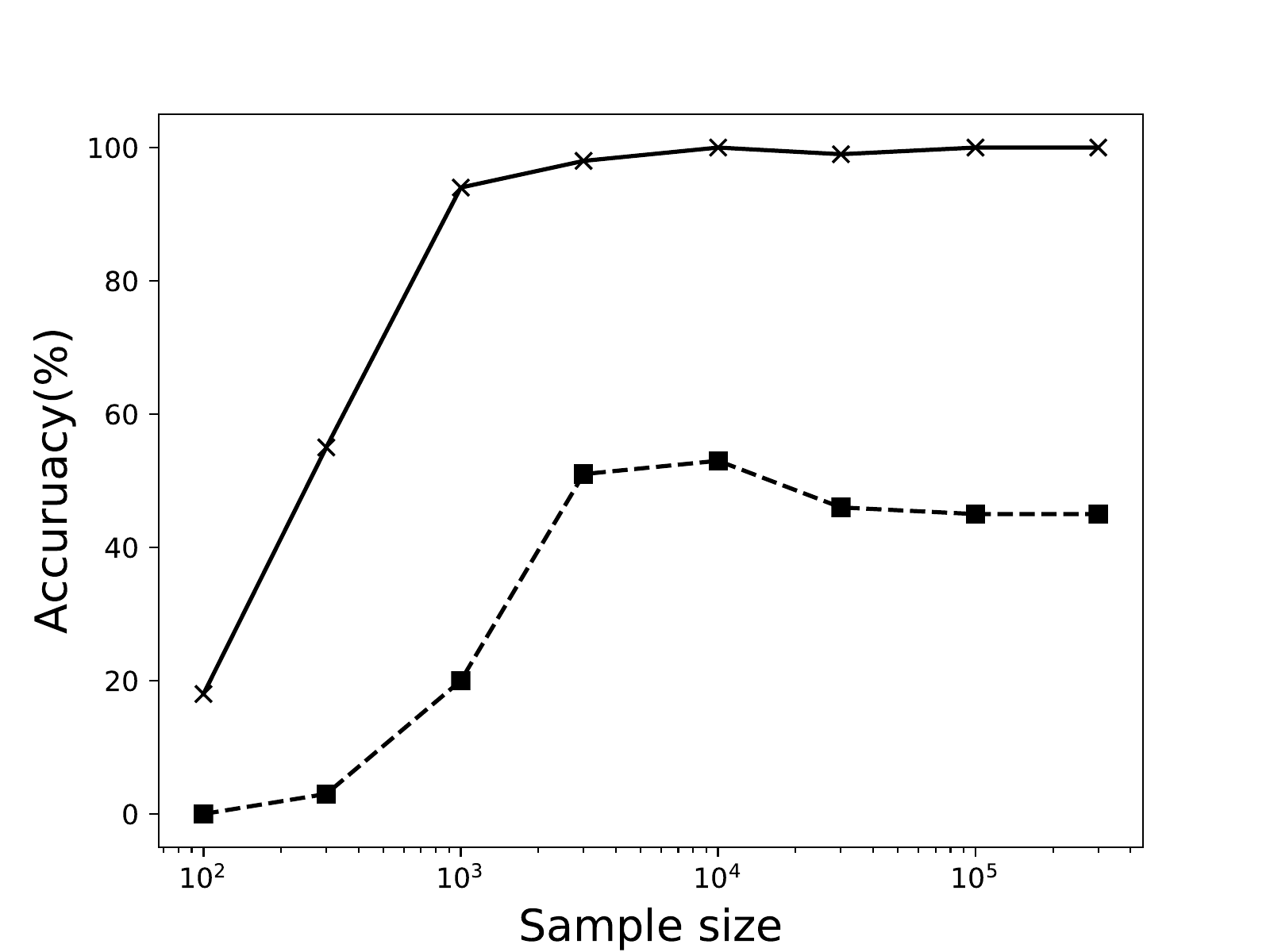}%
  }%
  \hfill
  \subfigure[c=3]{%
    \includegraphics[width=0.33\textwidth]{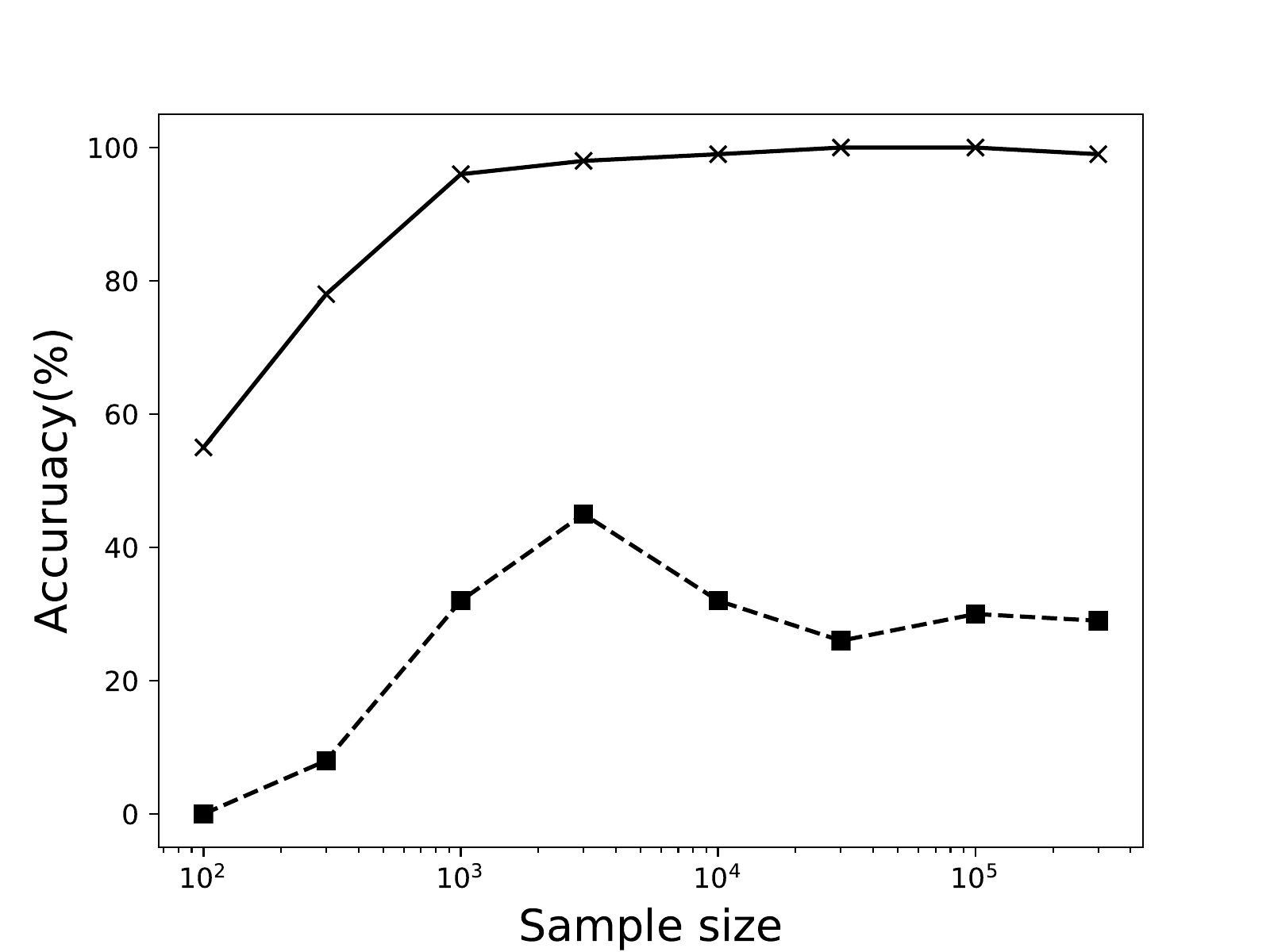}%
  }%
  \caption{The accuracy rate of estimating skeleton from mixed continuous and discrete data for increasing number of samples and the number $c \in \{1, 2, 3\}$ of continuous variables involved.}
  \label{fig:c123}
\end{figure*}

The simulation study was conducted on a set of random DAG models with both continuous and discrete variables.
Each random graph was generated by adding an edge with probability 0.5 for each pair of possible nodes under constraints that the newly added edge will not introduce any directed cycle.
In the data generating process, each continuous variable $X_i$ is caused by the function $X_i =b_{i0} + e_i + \sum_{j \in \mathbf{PA}(i) }{b_{ij} X_j}$; each discrete variable $X_i$ was sampled from the probability distribution $P(X_i) \mid X_{\mathbf{PA}(i)}) = sigmoid(b_{i0} + \sum_{j \in \mathbf{PA}(i) }{b_{ij} X_j)}$.
In all results presented, parameters $b_{ij}$ were chosen uniformly at random in the range [-1, -0.5] or [0.5, 1].

Using the BIC scoring function in equation (\ref{eq:bic}), we searched for the structure with maximum score among all possible structures.
Our search algorithm is implemented in Python 3.
We compared our algorithm against the PC algorithm which is implemented in the latest versions of the pgmpy library.
Since the discrete PC algorithm can not directly applied on the mixed continuous and discrete data, we discretized all the continuous data using mean value~\cite{YuanM13}. 
Since the PC algorithm does not recover all directions of the DAG, we only measure how often the PC algorithm can correctly infer the skeleton of the true DAG, which is the undirected graph resulting from removing all arrowheads from the DAG.

Procedures used for the simulation experiments are described below.
\begin{enumerate}
  \item For different number of continuous variables $c \in \{1,2,3\}$, we simulated 100 random DAGs with 4 variables. In total, we generated 300 random graphs.

  \item Using BIC score as we described previously, the causal structures were estimated based on $n \in \{100, 300, 1000, 3000,10000, 30000, 100000, 300000\}$ samples, respectively.

  \item The accuracy rate of recovering causal structure was calculated based on 100 iterations.
\end{enumerate}

Figure~\ref{fig:c123} provides a comparison of our proposed algorithm and the PC algorithm in terms of skeleton accuracy.
First, we observed that our proposed method learns the correct causal structure as the number of data increases.
The results empirically show that our method has the asymptotic consistency.
In contrast, the discrete PC algorithm was not able to estimate the correct causal structure even for large numbers of samples.
Second, as the number of continuous variables increases, the performance of the PC algorithm decreases.
The reason is considered that some information might be lost due to discretization.
The more number of continuous variables are discretized, the more information loses.

\begin{figure}[!ht]
  \begin{center}
    \includegraphics[width=0.33\textwidth]{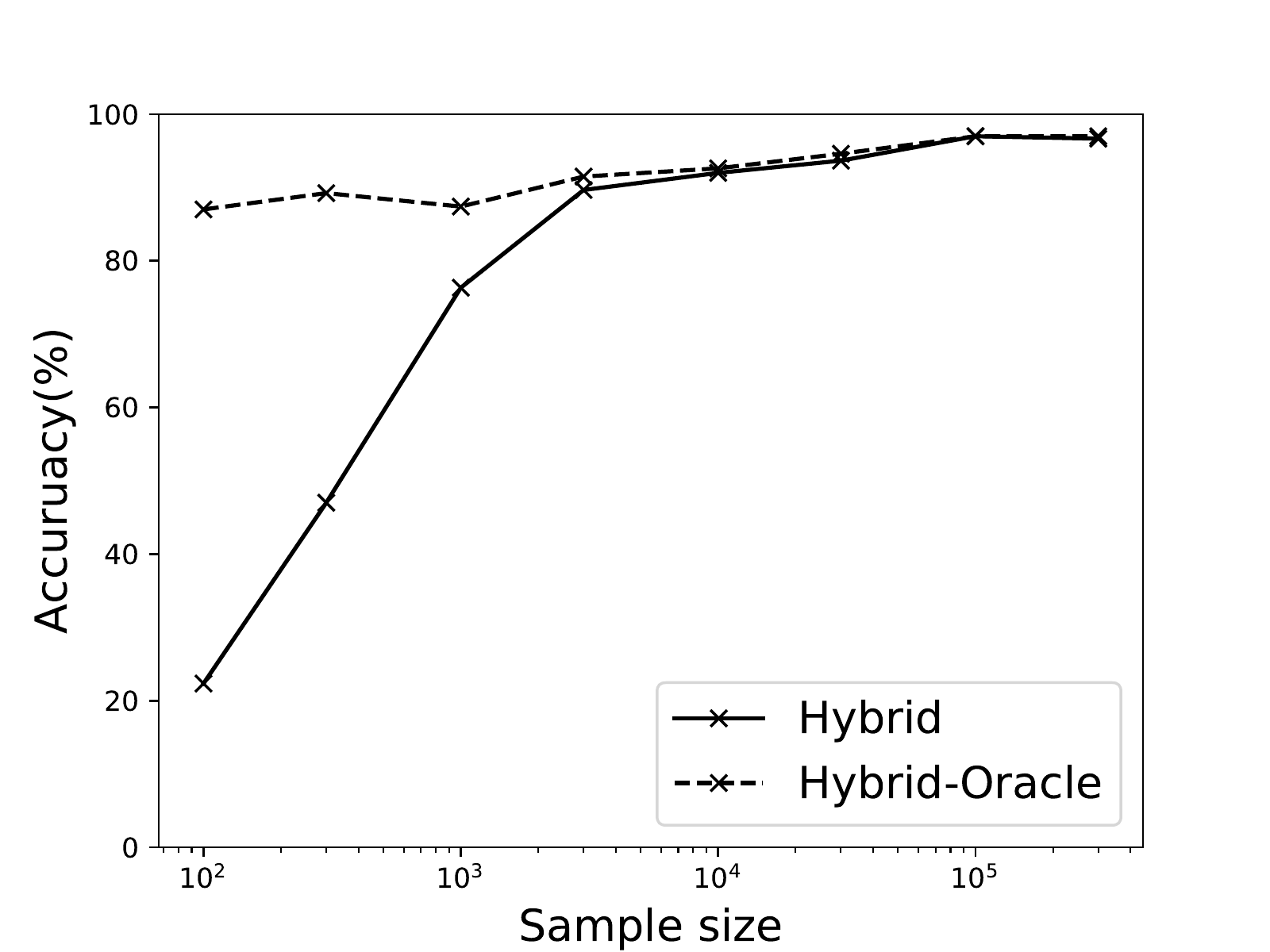}
  \end{center}
  \caption{A comparison of the hybrid algorithm and the hybrid-oracle algorithm with respect to the accuracy rate of estimating the DAG structure}
  \label{fig:oracle}
\end{figure}

In another experiment, we assume that the skeleton of the DAG model can be obtained from some oracle procedure.
Then we score each DAG which is consistent with the skeleton.
Figure~\ref{fig:oracle} reported the accuracy rate of this hybrid-oracle approach on the previous generated 300 DAG models.
In comparison, we also reported the performance analysis of full search approach.
A key observation is that when we know the undirected structure, the hybrid-oracle can learn full causal structure with $80\%$ accuracy, even for only 100 data samples.

This insight is quite important in practical case.
Because in many analysis we might know some pair of variables are correlate but do not know the causal direction.
Our BIC score and search approach can get efficiency if we know the undirected structure.
On they other hand, if we start from undirected structure instead of scratch, the necessary number of data samples decreases, which is a huge advantage for practice.

\section{Summary}  \label{sec:conclude}

Causal discovery from a mixture of continuous and discrete data is a important research problem and has practical value.
Most existing causal discovery methods either ignore the discrete data or discretize all the continuous data.
In this paper we proposed a hybrid causal model and derived the BIC scoring function for evaluating our model.






\bibliographystyle{named}
\bibliography{BibTeX/chaoli}

\begin{thebibliography}{}

\bibitem[\protect\citeauthoryear{Bishop}{2006}]{Bishop:2006}
Christopher~M. Bishop.
\newblock {\em Pattern Recognition and Machine Learning}.
\newblock Springer, 2006.

\bibitem[\protect\citeauthoryear{Chen \bgroup \em et al.\egroup
  }{2017}]{Chen2017}
Yi-Chun Chen, Tim~A. Wheeler, and Mykel~J. Kochenderfer.
\newblock Learning discrete {B}ayesian networks from continuous data.
\newblock {\em Journal of Artificial Intelligence Research}, 59:103--132, 2017.

\bibitem[\protect\citeauthoryear{Geiger and Heckerman}{1994}]{Geiger1994}
Dan Geiger and David Heckerman.
\newblock Learning gaussian networks.
\newblock In {\em Proceedings of the Tenth Conference on Uncertainty in
  Artificial Intelligence (UAI 1994)}, pages 235--243. Morgan Kaufmann, 1994.

\bibitem[\protect\citeauthoryear{Hoyer \bgroup \em et al.\egroup
  }{2008}]{Hoyer08}
Patrik~O. Hoyer, Aapo Hyv{\"{a}}rinen, Richard Scheines, Peter Spirtes, Joseph
  Ramsey, Gustavo Lacerda, and Shohei Shimizu.
\newblock Causal discovery of linear acyclic models with arbitrary
  distributions.
\newblock In {\em Proceedings of the 24th Conference in Uncertainty in
  Artificial Intelligence (UAI 2008)}, pages 282--289. AUAI Press, 2008.

\bibitem[\protect\citeauthoryear{Hyv{\"{a}}rinen \bgroup \em et al.\egroup
  }{2010}]{Aapo2010}
Aapo Hyv{\"{a}}rinen, Kun Zhang, Shohei Shimizu, and Patrik~O. Hoyer.
\newblock Estimation of a structural vector autoregression model using
  non-gaussianity.
\newblock {\em Journal of Machine Learning Research}, 2010.

\bibitem[\protect\citeauthoryear{Koller and
  Friedman}{2009}]{Koller+Friedman:09}
D.~Koller and N.~Friedman.
\newblock {\em Probabilistic Graphical Models: Principles and Techniques}.
\newblock MIT Press, 2009.

\bibitem[\protect\citeauthoryear{Morgan and Winship}{2007}]{Morgan2007}
Stephen~L. Morgan and Christopher Winship.
\newblock {\em Counterfactuals and Causal Inference: Methods and Principles for
  Social Research}.
\newblock Cambridge University Press, 2007.

\bibitem[\protect\citeauthoryear{Park and Raskutti}{2015}]{ParkR15}
Gunwoong Park and Garvesh Raskutti.
\newblock Learning large-scale poisson {DAG} models based on overdispersion
  scoring.
\newblock In {\em {NIPS}}, pages 631--639, 2015.

\bibitem[\protect\citeauthoryear{Pearl}{2000}]{Pearl2000}
Judea Pearl.
\newblock {\em Causality: Models, Reasoning, and Inference}.
\newblock Cambridge University Press, 2000.

\bibitem[\protect\citeauthoryear{Peters \bgroup \em et al.\egroup
  }{2011}]{Peters11}
J.~Peters, D.~Janzing, and B.~Scholkopf.
\newblock Causal inference on discrete data using additive noise models.
\newblock {\em IEEE Transactions on Pattern Analysis and Machine Intelligence},
  33(12):2436--2450, 2011.

\bibitem[\protect\citeauthoryear{Schwarz}{1978}]{schwarz1978}
Gideon Schwarz.
\newblock Estimating the dimension of a model.
\newblock {\em The Annals of Statistics}, 6(2):461--464, 1978.

\bibitem[\protect\citeauthoryear{Shimizu \bgroup \em et al.\egroup
  }{2006}]{shimizu06a}
Shohei Shimizu, Patrik~O. Hoyer, Aapo Hyv{\"a}rinen, and Antti Kerminen.
\newblock A linear non-gaussian acyclic model for causal discovery.
\newblock {\em Journal of Machine Learning Research}, 7:2003--2030, 2006.

\bibitem[\protect\citeauthoryear{Spirtes \bgroup \em et al.\egroup
  }{1993}]{Spirtes1993}
Peter Spirtes, Clark Glymour, and Richard Scheines.
\newblock {\em Causation, Prediction, and Search}.
\newblock Springer-Verlag New York, 1993.

\bibitem[\protect\citeauthoryear{Yehezkel and Lerner}{2009}]{Yehezkel2009}
Raanan Yehezkel and Boaz Lerner.
\newblock Bayesian network structure learning by recursive autonomy
  identification.
\newblock {\em Journal of Machine Learning Research}, 10:1527--1570, 2009.

\bibitem[\protect\citeauthoryear{Yuan and Malone}{2013}]{YuanM13}
Changhe Yuan and Brandon~M. Malone.
\newblock Learning optimal {B}ayesian networks: A shortest path perspective.
\newblock {\em Journal of Artificial Intelligence Research}, 48:23--65, 2013.

\end{thebibliography}
\end{document}